# Comparative Analysis of Object Detection Algorithms for Surface Defect Detection

Arpan Maity[1], Tamal Ghosh[2*]

[1,2] Department of Computer Science and Engineering, Adamas University
AKC Campus, Barrackpore-Barasat Road, P.O. Barbaria, PIN 700126, Barasat, West Bengal
Email: arpan1.maity@stu.adamasuniversity.ac.in [1], tamal.ghosh1@adamasuniversity.ac.in [2]
[*]Corresponding Author (Ph. +91-9163253061)

*Abstract*

This article compares the performance of six prominent object detection algorithms YOLOv11, RetinaNet, Fast R-CNN, YOLOv8, RT-DETR, and DETR on the NEU-DET surface defect detection dataset comprising images representing various metal surface defects, a crucial application in industrial quality control. Each model's performance was assessed regarding detection accuracy, speed, and robustness across different defect types such as scratches, inclusions, and rolled-in scales. YOLOv11, a state-of-the-art real-time object detection algorithm, demonstrated superior performance compared to the other methods, achieving a remarkable 70% higher accuracy on average. This improvement can be attributed to YOLOv11's enhanced feature extraction capabilities and ability to process the entire image in a single forward pass, making it faster and more efficient in detecting smaller surface defects. Additionally, YOLOv11's architecture optimizations, such as improved anchor box generation and deeper convolutional layers, contributed to more precise localization of defects. In conclusion, YOLOv11's outstanding performance in accuracy and speed solidifies its position as the most effective model for surface defect detection on the NEU dataset, surpassing competing algorithms by a substantial margin.

## 1 INTRODUCTION

Metal planar materials such as steel, aluminium, and copper play a crucial role in a wide array of industries, including automotive, aerospace, and manufacturing. In particular, within Micro, Small, and Medium Enterprises (MSMEs), which form the backbone of many manufacturing sectors, ensuring the quality of metal surfaces is vital for maintaining competitiveness, reducing waste, and minimizing costs (Candraningrat et al., 2021). Surface defects, such as dents, cracks, and corrosion, if left undetected, can lead to serious issues in product performance, customer dissatisfaction, and economic losses. For MSMEs, which often operate with constrained resources and tight budgets, efficient surface defect detection can significantly impact their operational efficiency and profitability (Sharma, 2021). Traditionally, the task of surface defect detection has been carried out manually by trained human inspectors. However, this process is time-consuming, error-prone, and highly dependent on the experience of the inspector. Additionally, visual fatigue can result in missed defects, leading to unreliable quality assurance. The need for automated, cost-effective, and robust surface defect detection systems has become increasingly apparent, especially in the MSME sector, where production downtime and high labour costs can be detrimental (Chen et al., 2021). In response to this need, computer vision-based Automated Visual Inspection Systems (AVIS) have gained prominence for their ability to enhance the speed and accuracy of defect detection. These systems, powered by machine learning and deep learning models, have the potential to provide real-time, consistent, and scalable inspection solutions (Rippel & Merhof, 2023). This study proposes a comparative analysis of anchor-based and anchor-free object detection algorithms for surface defect detection using the NEU-DET dataset (Song & Yan, 2013), which consists of grayscale images of steel surfaces with six defect categories. The goal is to evaluate the performance of both anchor-based methods, such as YOLOv11 (Jiang et al., 2022), Faster R-CNN (Ren et al., 2015), and RetinaNet (Li & Ren, 2019), and anchor-free methods, including RT-DETR (M. Zhu & Kong, 2024), YOLOv8 (G. Wang et al., 2023), DETR (X. Zhu et al., 2020). The NEU-DET dataset, although widely used in surface defect detection research, poses several challenges due to its limited dataset size and grayscale nature, making it an ideal candidate for evaluating the performance of these algorithms in real-world MSME settings. The contributions of this study are as follows:

1. A detailed comparison of anchor-based and anchor-free object detection algorithms on the NEU-DET dataset.
2. Evaluation of the performance of these models in detecting surface defects common in MSMEs, with a focus on precision and overall model efficiency.
3. Examination of the trade-offs between model accuracy, computational complexity, and deployment feasibility in resource-constrained MSME environments.

This paper is organized in the following manner, Section #2 discusses related works in surface defect detection and object detection models, Section #3 details the methodology and implementation of the object detection models used in this study, Section #4 presents the experimental results and analysis, and Section #5 provides the conclusions and future work.

## 2 RELATED WORKS

### 2.1 Datasets

In this section, an overview of commonly used steel surface defect datasets for defect detection in industrial settings is discussed. The Northeastern University (NEU-DET) as displayed in figure 1, surface defect dataset comprises 1,800 grayscale images categorized into six distinct types of steel surface defects. These categories include rolled-in scale (Rs), patches (P), crazing (Cr), pitted surface (Ps), inclusion (In), and scratches (Sc), with each defect type containing 300 images. The dataset highlights the variability within each defect category, as well as similarities between different defect categories. For example, within the patches (P) and scratches (Sc) categories, there are noticeable variations in appearance and texture, while rolled-in scale (Rs) and crazing (Cr) share some visual similarities.

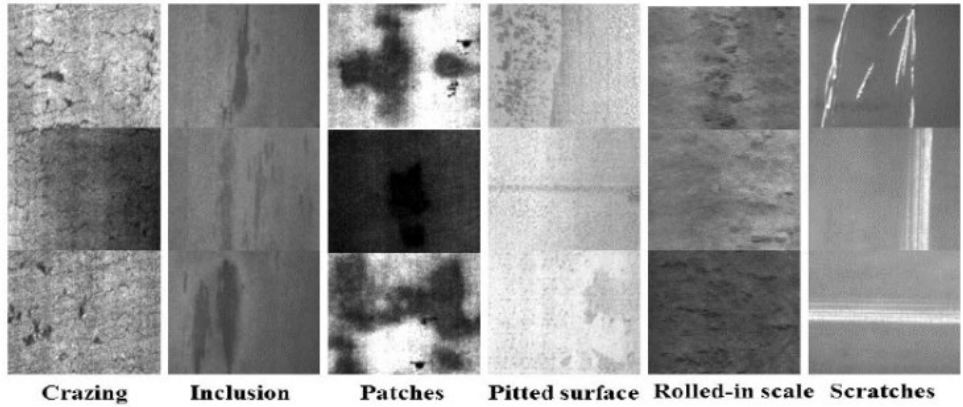

Figure 1. The NEU-DET defect dataset (Song & Yan, 2013)

This dataset is widely used in research for evaluating the performance of defect detection models due to its balanced class distribution and the diversity of defects.

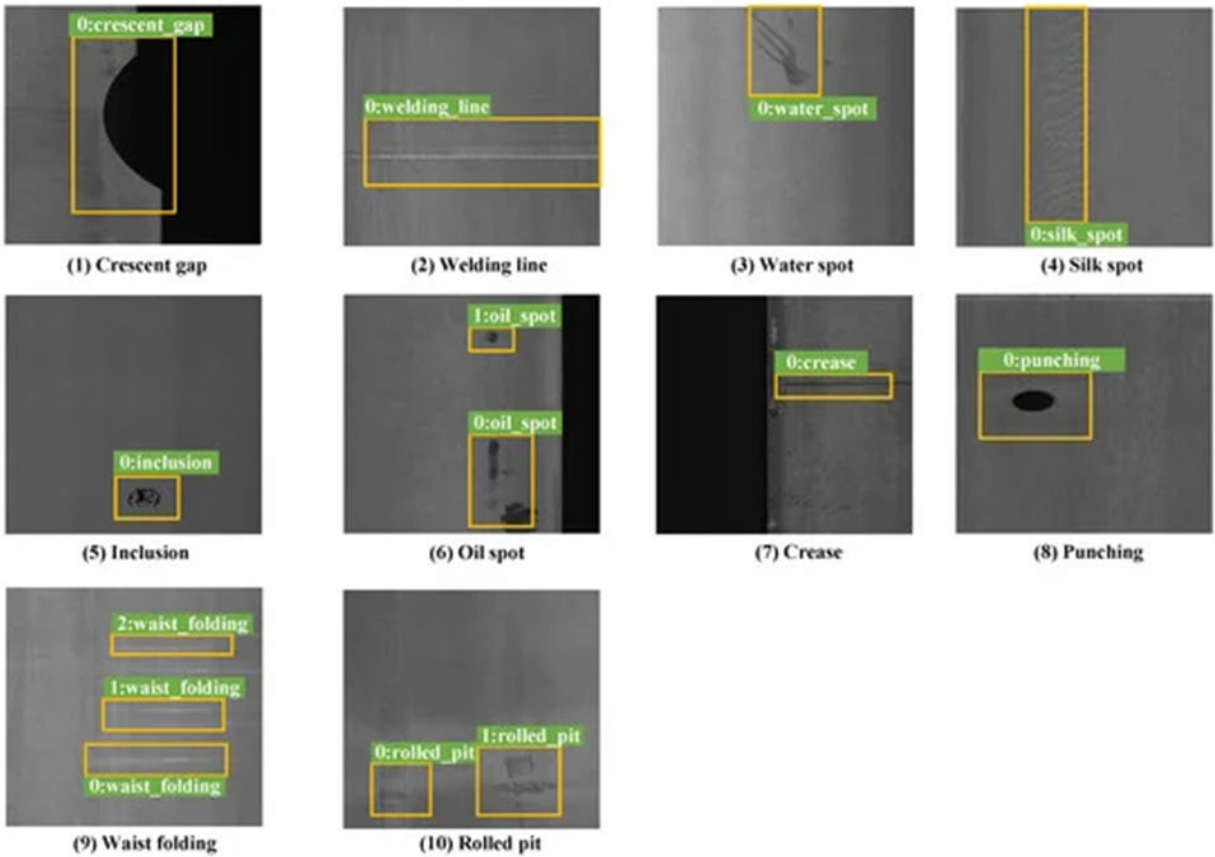

Figure 2. The GC10-DET defect dataset (Lv et al., 2020)

The GC10-DET dataset significantly advances surface defect detection datasets, offering 3,570 grayscale images sourced from real-world industrial steel plates. This dataset contains ten distinct categories of surface defects: punching, weld line, crescent gap, water spot, oil spot, silk spot, inclusion, rolled pit, crease, and waist folding. The larger size and greater variety of defect types make this dataset more suitable for developing robust machine-learning models capable of accurately detecting and classifying different surface defects. The diversity of defects and the dataset's real-world origins make it highly valuable for improving the practical applicability of defect identification tools in industrial settings. For this research, we perform our comparative analysis of object detection models on the NEU-DET dataset. The dataset's balanced class distribution and a substantial range of defect types provide a suitable benchmark for the performance assessment of both anchor-based and anchor-free methods in detecting and classifying surface defects.

## 2.2 Inspection Methods

Surface defect detection encompasses a wide array of methods that apply generic object detection techniques to identify predefined defect classes within an image, determining the spatial position and size of each defect instance. These techniques typically use bounding boxes to localize detected defects and can be categorized as either anchor-based or anchor-free object detectors. Below, we present a concise review of the selected methods used in this research.

*2.2.1 Anchor-Based Detectors*

Anchor-based object detection methods rely on predefined proposals (anchors) to identify objects by dividing the image into regions, followed by bounding box refinement to improve accuracy. These detectors can be grouped into two types: two-stage detectors and one-stage detectors. Two-stage detectors, such as faster R-CNN, first generate object proposals before predicting object classes and refining bounding boxes in the second stage. One-stage detectors, such as YOLOv11 and RetinaNet, predict bounding boxes and object classes in a single pass through the network, eliminating the need for a proposal generation stage. Faster R-CNN is a classic two-stage detector that separates proposal generation and classification. While it delivers high accuracy, it often involves a more computationally intensive process due to its two-stage approach (Ren et al., 2015). YOLOv11 is a more advanced version of the You Only Look Once (YOLO) family, a one-stage detector known for its real-time performance and high detection speed. YOLOv11 leverages anchor boxes to predict object locations and class labels simultaneously, making it efficient for detecting defects in real-time industrial environments (Jiang et al., 2022). RetinaNet also falls under the one-stage detector category, offering a balance between speed and accuracy. It improves detection by using a specialized focal loss function, which helps mitigate the imbalance between foreground and background classes in defect detection (Li & Ren, 2019). However, one limitation of anchor-based detectors is the need for extensive hyperparameter tuning to adjust anchor sizes, aspect ratios, and scales to achieve optimal performance, particularly in detecting small or irregularly shaped surface defects (Yang et al., 2024).

*2.2.2 Anchor-Free Detectors*

In contrast to anchor-based methods, anchor-free detectors eliminate the need for predefined anchors by directly identifying objects based on learned features, making the detection process more flexible. These methods are generally divided into key point-based and centre-based approaches. YOLOv8 is an advanced iteration in the YOLO series that incorporates an anchor-free design. It uses keypoint-based detection methods and avoids anchor boxes altogether by directly predicting the centre of the object and the dimensions of the bounding box. YOLOv8 offers a significant improvement in detection accuracy and speed, particularly for detecting smaller or irregularly shaped defects on industrial metal surfaces (G. Wang et al., 2023). RT-DETRT (Real-Time DEtection Transformer) leverages transformers for object detection and focuses on direct detection without anchor boxes, relying on attention mechanisms to identify and localize defects (M. Zhu & Kong, 2024). This anchor-free method offers the advantage of simplicity and flexibility in handling various defect shapes and sizes. DETR (Detection Transformer) is a fully transformer-based object detection model that represents a significant shift from traditional anchor-based methods. DETR directly predicts object locations and classifications without using predefined anchor boxes. It leverages an attention mechanism to learn spatial relationships within the image, making it highly effective for surface defect detection, especially in complex and cluttered environments where defects may vary significantly in size and shape.

**3 METHODOLOGIES**

In this study, we conducted a comprehensive comparative analysis of various state-of-the-art object detection algorithms to assess their effectiveness in detecting common surface defects in metal surfaces, including crazing, inclusion, patches, pitted surfaces, rolled-in scale, and scratches. Our focus encompasses both anchor-based and anchor-free object detectors, selecting models that are popular in the literature for their advancements in object detection. The architectures of the algorithms are portrayed in figure 3-6.

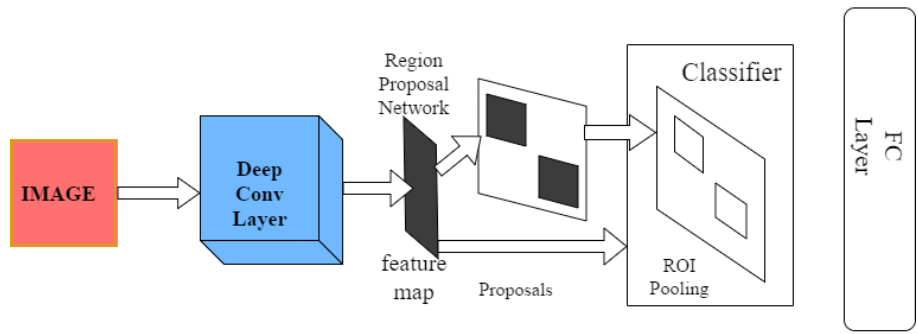

Figure 3. Faster RCNN architecture (Jha et al., 2019)

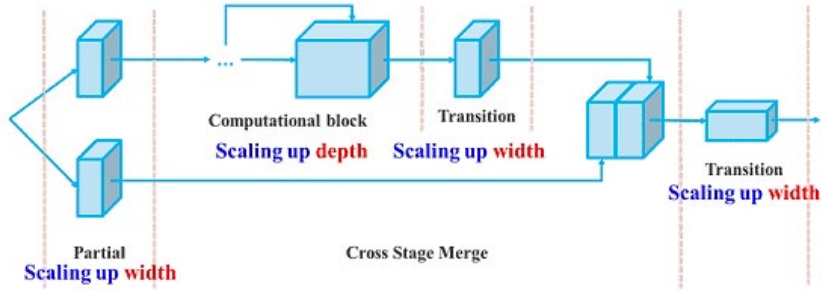

Figure 4. YOLO architecture (C.-Y. Wang et al., 2023)

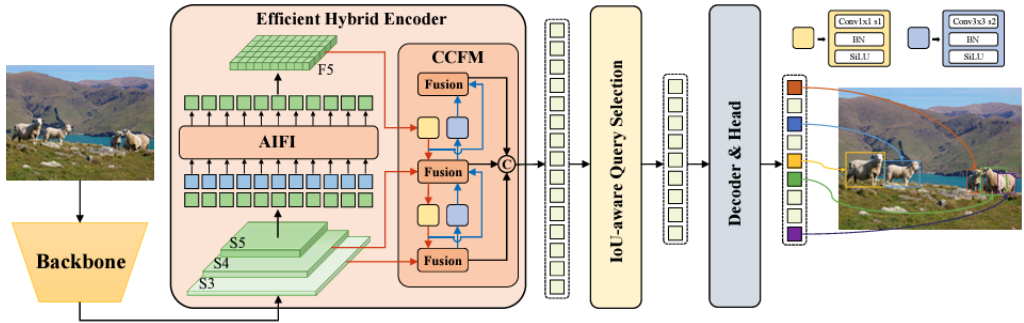

Figure 5. RT-DETR architecture (M. Zhu & Kong, 2024)

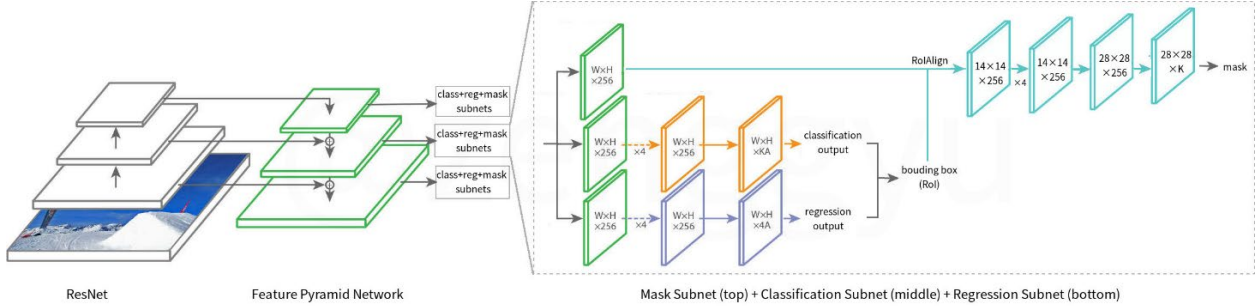

Figure 6. RetinaNet architecture (Li & Ren, 2019)

### 3.1 Implementation details

The experiments for this study were conducted using PyTorch, alongside Detectron2 (Pham et al., 2020) and YOLO. Three CUDA 12.4-enabled GPUs are utilized on a Google Colab environment to train all defect detection models, using the NEU-DET dataset as our benchmark. To ensure a fair comparison among the models, specific training configurations are established. For the Detectron2 models, specifically Faster R-CNN and RetinaNet were trained for 500 iterations, while the YOLO model was trained for 63 epochs. To calculate the number of iterations per epoch, a total of 1,260 training examples were considered with a batch size of 16, resulting in approximately 78.75 iterations per epoch. Therefore, the equivalent number of epochs for the 500 iterations of the Detectron2 models can be calculated as approximately 63 epochs. Additionally, The dataset was split into 70% for training, 20% for validation, and 10% for testing to facilitate an effective evaluation of the models. The implementation details of the seven object detection models are further explored,

*RetinaNET* (Li & Ren, 2019)
RetinaNet leverages a Cascade R-CNN architecture with a ResNet-50 backbone (Koonce, 2021) consisting of four stages, where the initial stage is frozen to facilitate feature extraction. During training, the model employs batch normalization in evaluation mode and is initialized with pretrained ResNet-50 weights from Detectron2. For classification, it uses focal loss combined with sigmoid activation, while bounding box regression is handled using L1 loss. The optimization process is conducted via Stochastic Gradient Descent (SGD) (Bottou, 2012) with a learning rate of 0.0025, momentum set to 0.9, and a weight decay of 0.0001 with 500 iterations.

*Faster R-CNN* (Ren et al., 2015)
The Faster R-CNN model leverages the faster_rcnn_R_50_FPN_3x backbone (Ren et al., 2015) weights from Detectron2, structured with four stages and incorporating batch normalization. It uses a Feature Pyramid Network (FPN) (Lin et al., 2016), which extracts features from four different layers—256, 512, 1024, and 2048—producing five output stages. The input image resolution for this configuration is 640×640 pixels. For training, the model employs cross-entropy loss for classification and L1 loss for bounding box regression, with equal weighting given to both losses. The optimization process is carried out using Stochastic Gradient Descent (SGD) (Bottou, 2012), with a learning rate of 0.0025, a momentum of 0.9, and a weight decay of 0.0001. The learning rate schedule follows a step policy, with 500 iterations.

*YOLOv11* (Jiang et al., 2022)
The YOLOv11 model is trained using the pre-trained weights model=yolo11s, which are specifically fine-tuned for object detection tasks on 640×640 resolution images. This model configuration is capable of identifying small surface defects with high accuracy and gives real-time performance. The backbone utilizes the CSP-Darknet architecture (Misra, 2019), combining Convolutional Neural Networks (CNNs) (O'Shea & Nash, 2015) and Cross-Stage Partial (CSP) layers (C.-Y. Wang et al., 2021), which optimize feature extraction while minimizing computational costs. The training process leverages binary cross-entropy loss for classification and CIoU (Complete Intersection over Union) loss for bounding box regression, ensuring precise object localization. Optimization is conducted using Stochastic Gradient Descent (SGD) (Bottou, 2012) with a learning rate of 0.01, momentum of 0.937, and a weight decay of 0.0005.

*RTDETR* (M. Zhu & Kong, 2024)

The RT-DETR (Real-Time Detection Transformer) model is trained using the pre-trained weights model. The model leverages the Transformer-based architecture (Y. Wang et al., 2022), which uses multi-head self-attention mechanisms to capture long-range dependencies and relationships between features, providing an advantage in detecting complex surface defects over large areas. For classification, the model utilizes cross-entropy loss, while L1 loss is applied for bounding box regression, ensuring precise localization of surface defects. The optimization is performed using AdamW optimizer (Llugsi et al., 2021), with a learning rate of 0.0001, weight decay of 0.01, and a momentum of 0.9. The model employs a cosine decay learning rate schedule, ensuring smooth convergence and improved performance, with warm-up steps for the first 300 iterations to stabilize the training process.

*YOLOv8* (G. Wang et al., 2023)
The YOLOv8 model is trained using the pre-trained weights model=yolov8s.pt (G. Wang et al., 2023), is purposely designed to identify surface defects on 640×640 resolution images. YOLOv8 builds on the strong foundation of previous YOLO models, providing improved accuracy, efficiency, and speed in object detection. The model incorporates an advanced CSP-Darknet backbone (Misra, 2019), along with a Path Aggregation Network (PAN) (Yu et al., 2022) for enhancing feature propagation, and Spatial Pyramid Pooling (SPP) (He et al., n.d.; Yu et al., 2022) for better handling of multi-scale objects. For the training process, the model uses binary cross-entropy loss for classification and CIoU (Complete Intersection over Union) loss for bounding box regression. The optimization is performed using the AdamW optimizer (Llugsi et al., 2021) with a learning rate of 0.001, a momentum of 0.937, and a weight decay of 0.0005.

*DETR* (X. Zhu et al., 2020)
The DEtection TRansformers (DETR) model used is the ResNet-50 version (Koonce, 2021). This model leverages a transformer architecture for end-to-end object detection while not relying on the anchor boxes. The image resolution during training and testing is set to 640×640. The pre-trained model checkpoint is used, and the confidence threshold is set to 0.5, with an Intersection over Union (IoU) threshold of 0.8 to filter predictions. The DETR model is trained using cross-entropy loss for classification and L1 loss for bounding box regression. The AdamW optimizer (Llugsi et al., 2021) is employed, with a learning rate of 1e-4, weight decay of 0.0001, and the learning rate follows a step schedule, reducing at specific milestones. The input images undergo normalization, and the model employs a linear warm-up strategy at the start of training for 500 iterations.

## 3.2 Performance evaluation metrics

To assess the performance of different methods on the NEU-DET dataset for surface defect detection, we utilized the Average Precision (AP) metric, as outlined in Equation (3). This is computed using Equations (1) and (2), while varying the Intersection over Union (IoU) thresholds defined in Equation (4) and considering different defect scales. This approach allows us to evaluate the overall effectiveness of the various detection models.

To evaluate the effectiveness of various methods on the NEUDET dataset for surface defect detection, we employ the Average Precision (AP) metric as described in Equation (3) (Misra, 2019). This metric is calculated using Equations (1) and (2), with different Intersection over Union (IoU) thresholds specified in Equation (4) and taking into account varying scales of defects. This methodology provides a comprehensive assessment of the performance of the detection models.

AP, or Average Precision, is the primary evaluation metric that quantifies performance across various Intersection over Union (IoU) thresholds, ranging from 0.5 to 0.95 in increments of 0.05. This metric assesses the algorithm's effectiveness at different IoU cutoffs, providing a comprehensive view of its detection capabilities.

AP50, or Average Precision at IoU 50%, is a specific variant of the Average Precision metric. It considers a detection correct if the Intersection over Union (IoU) between the predicted bounding box and the ground-truth bounding box exceeds 50%. This threshold assesses the model's accuracy in capturing surface defects while allowing for some degree of overlap (Abazović et al., 2024). While comparing the performance of different models, different metrics such as Average Precision (AP) and AP50 are used. In addition, to evaluate each model's performance different Average Precision for each individual class is used, which provides a more detailed comparison of how well each model performed across different object categories. This approach allowed for a comprehensive understanding of the strengths and weaknesses of each model in detecting specific classes.

AP (Crazing) is a metric that evaluates the model's precision in detecting crazing defects within the dataset. Crazing refers to the surface cracks or fissures found on metal surfaces, and this metric specifically indicates the model's effectiveness in accurately identifying such defects.

AP (Inclusion) measures the average precision for detecting inclusion defects, which are foreign materials embedded in the metal during the manufacturing process. This metric evaluates the model's effectiveness in accurately identifying these anomalies within the dataset.

AP (Patches) reflects the model's performance in detecting patch defects, which are irregular areas on the metal surface. This metric evaluates how accurately the model identifies these specific regions, providing insight into its capability to detect such anomalies.

AP (Pitted Surface) evaluates the model's capability to detect pitted surfaces, which are marked by small, irregular pits or indentations on the metal surface. This metric indicates how effectively the model recognizes these specific surface defects, providing insight into its performance in identifying pitting issues.

AP (Rolled-in Scale) measures the model's precision in detecting rolled-in scale defects, which occur when scale particles are embedded in the metal during rolling processes. This metric reflects the model's effectiveness in identifying this particular type of surface defect.

AP (Scratches) evaluates the model's performance in detecting scratches, a common type of surface defect caused by physical abrasions. This metric indicates the model's accuracy in identifying these defects, providing insights into its effectiveness in recognizing surface imperfections.

$$Precision = \frac{TP}{TP+FP} \tag{1}$$

$$Recall = \frac{TP}{TP+FN} \tag{2}$$

$$AP = \int_1^0 p(r)dr \tag{3}$$

$$IOU = \frac{Area\ of\ Overlap\ of\ bounding\ boxes}{Area\ of\ Union\ of\ bounding\ boxes} \tag{4}$$

Where, TP, FP, and FN denote the counts of true positives, false positives, and false negatives, respectively. The term p(r) refers to precision as a function of recall, indicating how the precision metric changes based on different levels of recall in the evaluation of detection performance.

## 4 RESULTS AND DISCUSSION

The findings presented in Table 1 assess the effectiveness of several object detection models, classified into anchor-based and anchor-free categories, specifically focusing on their performance in detecting surface defects. Average Precision (AP) serves as the primary metric for evaluation. In the overall AP comparison, YOLOv11, classified as an anchor-based model, stands out with the highest overall AP of 38.6%. In contrast, Faster R-CNN performs significantly inferior results, achieving only 9.7% AP.

Table 1. Comparison results of anchor-based and anchor-free methods

| | Anchor Based Method | | | Anchor Free Method | | |
|---|---|---|---|---|---|---|
| Metric | YOLO v11 | Faster RCNN | RetinaNet | RTDETR | YOLOv8 | DETR |
| Overall AP | 38.6% | 9.7% | 21.1% | 21.0% | 35.9% | 11.6% |
| AP@IOU=0.50 | 71.6% | 30.1% | 46.2% | 55.0% | 68.7% | 25.2% |
| AP(Crazing) | 21.3% | 8.0% | 12.8% | 18.2% | 17.6% | 8.2% |
| AP(Inclusion) | 44.1% | 2.9% | 21.1% | 27.6% | 39.7% | 14.3% |
| AP(Patches) | 60.5% | 17.3% | 47.3% | 14.9% | 57.4% | 6.9% |
| AP(Pitted Surface) | 43.8% | 11.8% | 31.7% | 31.4% | 43.4% | 17.8% |
| AP(Rolled-in Scale) | 23.0% | 11.5% | 13.9% | 19.7% | 22.3% | 10.1% |
| AP(Scratches) | 39.1% | 6.9% | 0.0% | 22.8% | 34.9% | 12.5% |

Among the anchor-free models, YOLOv8 delivers a commendable AP of 35.9%, closely following YOLOv11. Whereas, the transformer-based architecture of DETR under-performed, with just 11.6% AP, which indicates the requirement of further tuning for effective surface defect detection. Focusing on performance evaluated at an Intersection over Union (IOU) threshold of 0.50,

YOLOv11 excels with a notable AP of 71.6%, highlighting its exceptional ability for precise localization.

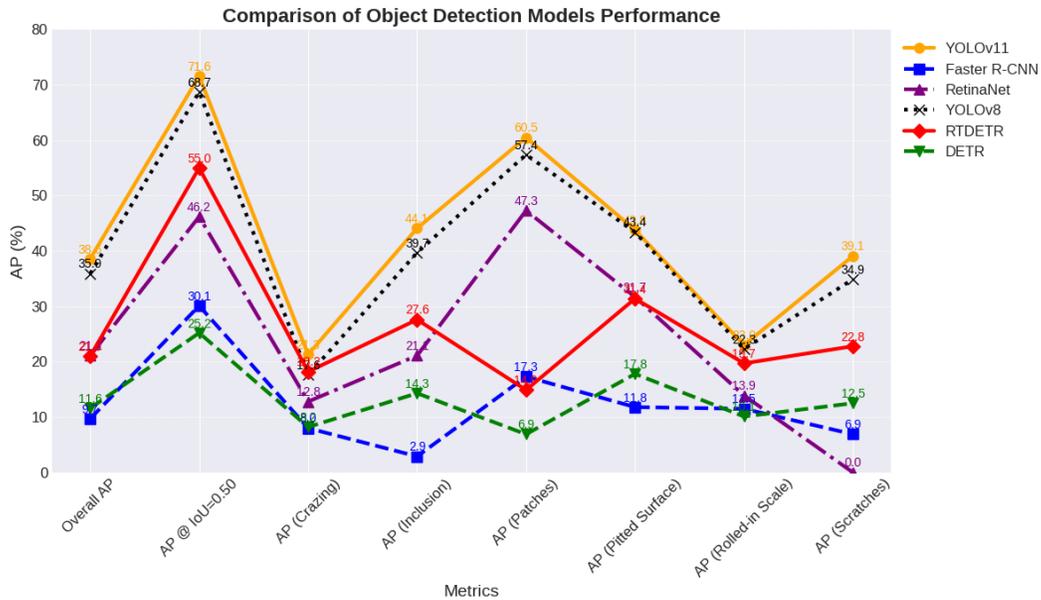

Figure 7. Graphical representation of the comparison results

YOLOv8 ranks second best with an AP of 68.7%, while DETR falls significantly short with an AP of 25.2%. Additionally, both RetinaNet and RT-DETR present competitive results at this IOU level, reporting APs of 46.2% and 55.0%, respectively. When examining class-specific performance, the models generally struggle with detecting crazing defects, as indicated by low AP values overall. YOLOv11 leads in this category with an AP of 21.3%, while Faster R-CNN and DETR show weaker performance, recording APs of 8.0% and 8.2%. In detecting inclusions, YOLOv11 and YOLOv8 show strong performance with APs of 44.1% and 39.7%. In contrast, DETR's ability to detect inclusions is limited, achieving only 14.3% AP, which may suggest challenges in identifying smaller or irregularly shaped defects. For patch detection, YOLOv11 again outpaces the rest with an AP of 60.5%, closely followed by YOLOv8 with 57.4% AP. RetinaNet provides reasonable results with an AP of 47.3%, while RT-DETR underperforms with just 14.9% AP. On the other hand pitted surfaces, YOLOv11 displays reliable detection with an AP of 43.8%, closely trailed by YOLOv8 at 43.4%. DETR again lags in this category, achieving only 17.8% AP, suggesting difficulties in recognizing smaller, textured defects. The detection of rolled-in scales presents lower performance across all models, with YOLOv11 again in the lead at 23.0% AP, while both YOLOv8 and RT-DETR perform similarly, reaching approximately 22.3% and 19.7%, respectively. Other models exhibit lower performance in this class. Finally, in the analysis of scratches, YOLOv11 continues to perform strongly with an AP of 39.1%, with RT-DETR achieving 22.8%. Meanwhile, Faster R-CNN and RetinaNet struggle considerably in this category, resulting in negligible or zero AP.

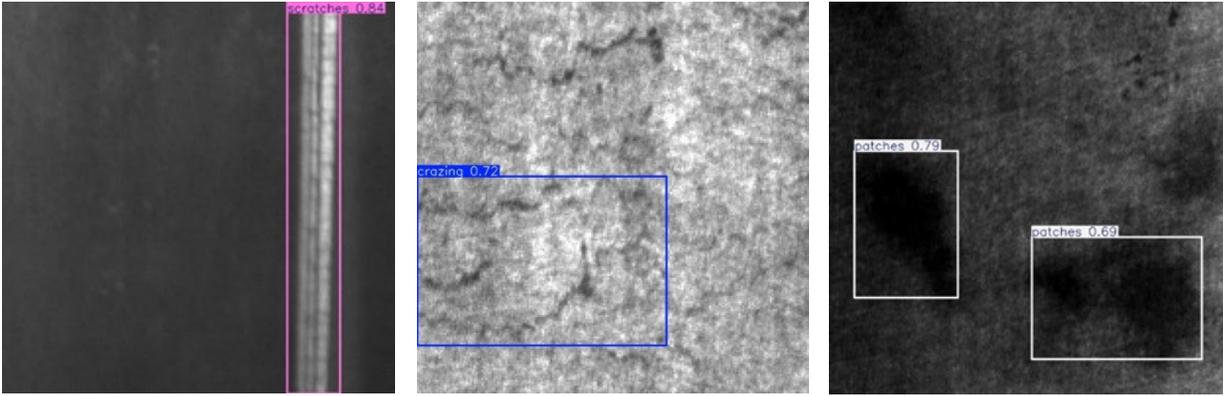

Figure 8. Sample visualization results of YOLOv11.

Overall, the anchor-based models, notably YOLOv11 and RetinaNet, consistently demonstrate exclusive performance when compared to DETR across various defect types. However, anchor-free models such as YOLOv8 show competitive capabilities with their anchor-based counterparts in many categories, occasionally surpassing them, as evidenced by high AP scores for inclusions, patches, and pitted surfaces. The sample outputs of YOLOv11 are displayed in Figure 8. Although DETR is recognized as a leading anchor-free approach, its inferior performance in complex defect detection indicates the need for further optimization for the effective identification of surface defects in metal planar materials.

## 4. Conclusions

While evaluating the surface defect detection models, anchor-based approaches such as YOLOv11 and RetinaNet have consistently outperformed others across various defect types, proving their robustness in real-world applications. However, the rise of anchor-free models, notably YOLOv8, presents a strong alternative, occasionally surpassing anchor-based models in detecting specific defects such as inclusions and pitted surfaces, with high AP scores. Despite the promise shown by anchor-free models, DETR, a prominent method in this category, struggles with complex surface defects, especially in metal planar materials. This suggests that while anchor-free models are advancing, they still require further refinement for such intricate tasks. The results, in this paper, emphasize the need for continued innovation, with hybrid models potentially clubbing the strengths of both anchor-based and anchor-free approaches to improve defect detection. As research progresses, further development will be critical to enhancing surface defect detection, particularly in quality control processes for manufacturing industries.

## 5. References

Abazović, A., Lekić, A., Jovović, I., Čakić, S., & Popović, T. (2024). Ovarian cancer detection using

computer vision. *2024 23rd International Symposium INFOTEH-JAHORINA (INFOTEH)*, *122*, 1–4. https://doi.org/10.1109/infoteh60418.2024.10495965

Bottou, L. (2012). Stochastic Gradient Descent Tricks. *Neural Networks: Tricks of the Trade*, 421–436. https://doi.org/10.1007/978-3-642-35289-8_25

Candraningrat, I., Abundanti, N., Mujiati, N., Erlangga, R., & Paper, I. (2021). The role of financial technology on development of MSMEs. *Accounting and Finance*, *7*(1), 225–230. http://m.growingscience.com/ac/Vol7/ac_2020_137.pdf

Chen, Y., Ding, Y., Zhao, F., Zhang, E., Wu, Z., & Shao, L. (2021). Surface defect detection methods for industrial products: A review. *Applied Sciences (Basel, Switzerland)*, *11*(16), 7657. https://doi.org/10.3390/app11167657

He, K., Zhang, X., Ren, S., & Sun, J. (n.d.). *Spatial Pyramid Pooling in Deep Convolutional Networks for Visual Recognition*. Retrieved October 17, 2024, from https://doi.org/10.1109/TPAMI.2015.2389824

Jha, S., Dey, A., Kumar, R., & Kumar-Solanki, V. (2019). A novel approach on visual question answering by parameter prediction using faster region based convolutional neural network. *International Journal of Interactive Multimedia and Artificial Intelligence*, *5*(5), 30. https://doi.org/10.9781/ijimai.2018.08.004

Jiang, P., Ergu, D., Liu, F., Cai, Y., & Ma, B. (2022). A Review of Yolo Algorithm Developments. *Procedia Computer Science*, *199*, 1066–1073. https://doi.org/10.1016/j.procs.2022.01.135

Koonce, B. (2021). ResNet 50. *Convolutional Neural Networks with Swift for Tensorflow*, 63–72. https://doi.org/10.1007/978-1-4842-6168-2_6

Li, Y., & Ren, F. (2019). Light-Weight RetinaNet for Object Detection. *ArXiv*. https://doi.org/10.48550/arXiv.1905.10011

Lin, T.-Y., Dollár, P., Girshick, R., He, K., Hariharan, B., & Belongie, S. (2016). *Feature Pyramid Networks for Object Detection*. http://arxiv.org/abs/1612.03144

Llugsi, R., Yacoubi, S. E., Fontaine, A., & Lupera, P. (2021, October 12). Comparison between Adam, AdaMax and Adam W optimizers to implement a Weather Forecast based on Neural Networks for the Andean city of Quito. *2021 IEEE Fifth Ecuador Technical Chapters Meeting (ETCM)*. 2021 IEEE Fifth Ecuador Technical Chapters Meeting (ETCM), Cuenca, Ecuador. https://doi.org/10.1109/etcm53643.2021.9590681

Lv, X., Duan, F., Jiang, J.-J., Fu, X., & Gan, L. (2020). Deep Metallic Surface Defect Detection: The New Benchmark and Detection Network. *Sensors*, *20*(6), 1562. https://doi.org/10.3390/s20061562

Misra, D. (2019). *Mish: A Self Regularized Non-Monotonic Activation Function*. http://arxiv.org/abs/1908.08681

O'Shea, K., & Nash, R. (2015). *An Introduction to Convolutional Neural Networks*. http://arxiv.org/abs/1511.08458

Pham, V., Pham, C., & Dang, T. (2020, December 10). Road damage detection and classification with Detectron2 and faster R-CNN. *2020 IEEE International Conference on Big Data (Big Data)*. 2020 IEEE International Conference on Big Data (Big Data), Atlanta, GA, USA. https://doi.org/10.1109/bigdata50022.2020.9378027

Ren, S., He, K., Girshick, R., & Sun, J. (2015). Faster R-CNN: Towards Real-Time Object Detection with Region Proposal Networks. *ArXiv*. https://doi.org/10.48550/arXiv.1506.01497

Rippel, O., & Merhof, D. (2023). Anomaly Detection for Automated Visual Inspection: A Review. *Bildverarbeitung in der Automation*, 1–13. https://doi.org/10.1007/978-3-662-66769-9_1

Sharma, R. K. (2021). *Quality management practices in MSME sectors*. Springer Singapore. https://doi.org/10.1007/978-981-15-9512-7

Song, K., & Yan, Y. (2013). A noise robust method based on completed local binary patterns for hot-rolled steel strip surface defects. *Applied Surface Science*, *285*, 858–864. https://doi.org/10.1016/j.apsusc.2013.09.002

Wang, C.-Y., Bochkovskiy, A., & Liao, H.-Y. M. (2021, June). Scaled-YOLOv4: Scaling cross stage partial network. *2021 IEEE/CVF Conference on Computer Vision and Pattern Recognition (CVPR)*. 2021

IEEE/CVF Conference on Computer Vision and Pattern Recognition (CVPR), Nashville, TN, USA. https://doi.org/10.1109/cvpr46437.2021.01283

Wang, C.-Y., Bochkovskiy, A., & Liao, H.-Y. M. (2023, June). YOLOv7: Trainable bag-of-freebies sets new state-of-the-art for real-time object detectors. *2023 IEEE/CVF Conference on Computer Vision and Pattern Recognition (CVPR)*. 2023 IEEE/CVF Conference on Computer Vision and Pattern Recognition (CVPR), Vancouver, BC, Canada. https://doi.org/10.1109/cvpr52729.2023.00721

Wang, G., Chen, Y., An, P., Hong, H., Hu, J., & Huang, T. (2023). UAV-YOLOv8: A Small-Object-Detection Model Based on Improved YOLOv8 for UAV Aerial Photography Scenarios. *Sensors*, *23*(16). https://doi.org/10.3390/s23167190

Wang, Y., Zhang, X., Yang, T., & Sun, J. (2022). Anchor DETR: Query Design for Transformer-Based Detector. *Proceedings of the AAAI Conference on Artificial Intelligence*, *36*(3), 2567–2575. https://doi.org/10.1609/aaai.v36i3.20158

Yang, Y., Yang, F., Sun, L., Wan, Y., & Lv, P. (2024). Anti-low angle resolution: Some adaptive improvements in anchor-based object detection algorithm for automotive radar target detection. *Digital Signal Processing*, *151*(104562), 104562. https://doi.org/10.1016/j.dsp.2024.104562

Yu, H., Li, X., Feng, Y., & Han, S. (2022). Multiple attentional path aggregation network for marine object detection. *Applied Intelligence*, *53*(2), 2434–2451. https://doi.org/10.1007/s10489-022-03622-0

Zhu, M., & Kong, E. (2024). Multi-Scale Fusion Uncrewed Aerial Vehicle Detection Based on RT-DETR. *Electronics*, *13*(8), 1489. https://doi.org/10.3390/electronics13081489

Zhu, X., Su, W., Lu, L., Li, B., Wang, X., & Dai, J. (2020). *Deformable DETR: Deformable Transformers for End-to-End Object Detection*. http://arxiv.org/abs/2010.04159